\documentclass[10pt,twocolumn,letterpaper]{article}

\usepackage[pagenumbers]{cvpr} 

\usepackage{graphicx}
\usepackage{amsmath}
\usepackage{amssymb}
\usepackage{booktabs}
\usepackage{appendix}
\usepackage{multirow}
\usepackage{graphicx}
\usepackage{makecell}
\usepackage{xcolor,colortbl}
\usepackage[accsupp]{axessibility}  
\usepackage[linesnumbered,ruled,vlined]{algorithm2e}

\usepackage[pagebackref,breaklinks,colorlinks]{hyperref}

\usepackage[capitalize]{cleveref}
\crefname{section}{Sec.}{Secs.}
\Crefname{section}{Section}{Sections}
\Crefname{table}{Table}{Tables}
\crefname{table}{Tab.}{Tabs.}
\crefname{equation}{Eq.}{Eqs.}
\crefname{algorithm}{Alg.}{Algs.}

\def\cvprPaperID{8321} 
\def\confName{CVPR}
\def\confYear{2023}

\begin{document}

\title{Architecture, Dataset and Model-Scale Agnostic Data-free Meta-Learning }

\author{
Zixuan Hu\textsuperscript{\rm 1}
\quad 
Li Shen\textsuperscript{\rm 2,}\thanks{Corresponding authors: Li Shen and Chun Yuan} 
\quad 
Zhenyi Wang\textsuperscript{\rm 3}
\quad 
Tongliang Liu\textsuperscript{\rm 4}
\quad 
Chun Yuan\textsuperscript{\rm 1,*}
\quad 
Dacheng Tao\textsuperscript{\rm 2}
\\
\textsuperscript{\rm 1}Tsinghua Shenzhen International Graduate School, China; 
\textsuperscript{\rm 2}JD Explore Academy, China \\
\textsuperscript{\rm 3}State University of New York at Buffalo, USA;
\textsuperscript{\rm 4}The University of Sydney, Australia\\
{\tt\small huzixuan21@mails.tsinghua.edu.cn; mathshenli@gmail.com; zhenyiwa@buffalo.edu}\\
{\tt\small tongliang.liu@sydney.edu.au; yuanc@sz.tsinghua.edu.cn; dacheng.tao@gmail.com
}
}

\maketitle

\begin{abstract}

The goal of data-free meta-learning is to learn useful prior knowledge from a collection of pre-trained models without accessing their training data. However, existing works only solve the problem in parameter space, which (i) ignore the fruitful data knowledge contained in the pre-trained models; (ii) can not scale to large-scale pre-trained models; (iii) can only meta-learn pre-trained models with the same network architecture. To address those issues, we propose a unified framework, dubbed \textbf{PURER}, which contains: (1) e\textbf{P}isode c\textbf{U}rriculum inve\textbf{R}sion (ECI) during data-free meta training; and (2) inv\textbf{E}rsion calib\textbf{R}ation following inner loop (ICFIL) during meta testing. 
During meta training, we propose ECI to perform pseudo episode training for learning to adapt fast to new unseen tasks. Specifically, we progressively synthesize a sequence of pseudo episodes by distilling the training data from each pre-trained model. The ECI adaptively increases the difficulty level of pseudo episodes according to the real-time feedback of the meta model. We formulate the optimization process of meta training with ECI as an adversarial form in an end-to-end manner. 
During meta testing, we further propose a simple plug-and-play supplement---ICFIL---only used during meta testing to narrow the gap between meta training and meta testing task distribution. Extensive experiments in various real-world scenarios show the superior performance of ours. 
Code is available at \url{https://github.com/Egg-Hu/PURER}.

\end{abstract}


\section{Introduction}

Meta-learning \cite{schmidhuber1987evolutionary,naik1992meta,bengio2013optimization} aims to learn useful prior knowledge (\eg, sensitive initialization) from a collection of similar tasks to facilitate the learning of new unseen tasks. Most meta-learning methods \cite{finn2017model,rajeswaran2019meta,bertinetto2018meta,wang2019bayesian,sun2021towards,bronskill2021memory,chen2021generalization,vanschoren2019meta,nichol2018first,hsu2018unsupervised,wang2022meta,wang2022learning} assume the access to the training and testing data of each task. However, this assumption is not always satisfied: many individuals and institutions only release the pre-trained models instead of the data. This is due to data privacy, safety, or ethical issues in real-world scenarios, making the task-specific data difficult or impossible to acquire. For example, many pre-trained models with arbitrary architectures are released on GitHub without training data. However, when facing a new task, we need some prior knowledge learned from those pre-trained models so that the model can be adapted fast to the new task with few labeled examples. Thus, meta-learning from several pre-trained models without data becomes a critical problem, named \textit{Data-free Meta-Learning (DFML)} \cite{wang2022metalearning}.

\begin{figure}[t]
  \centering
    \includegraphics[width=0.8\linewidth]{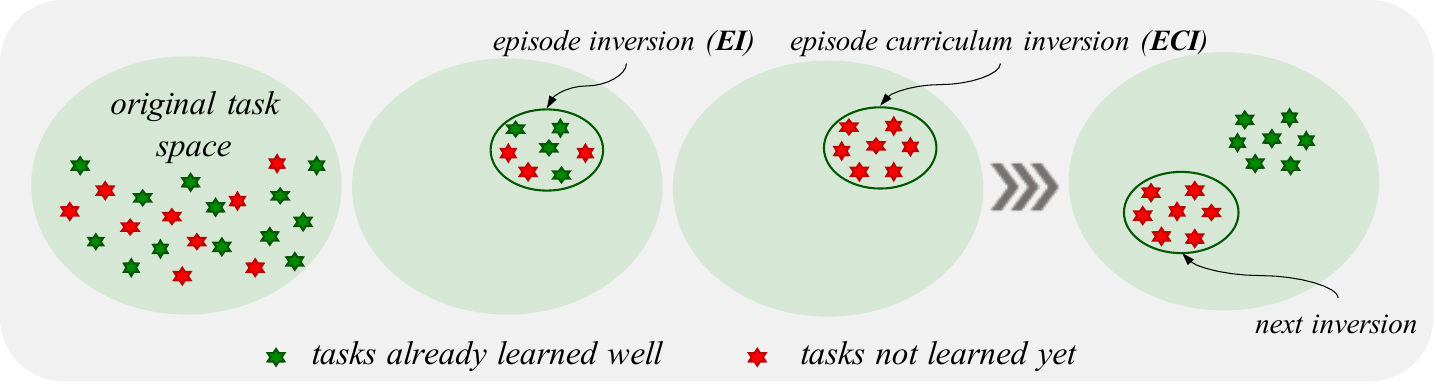}

   \caption{Episode Curriculum Inversion can improve the efficiency of pseudo episode training. At each episode, EI may repeatedly synthesize the tasks already learned well, while ECI only synthesizes harder tasks not learned yet.}
   \label{fig:curriculum}
   \vspace{-0.4cm}
\end{figure}

\begin{figure}[t]
  \centering
    \includegraphics[width=0.8\linewidth]{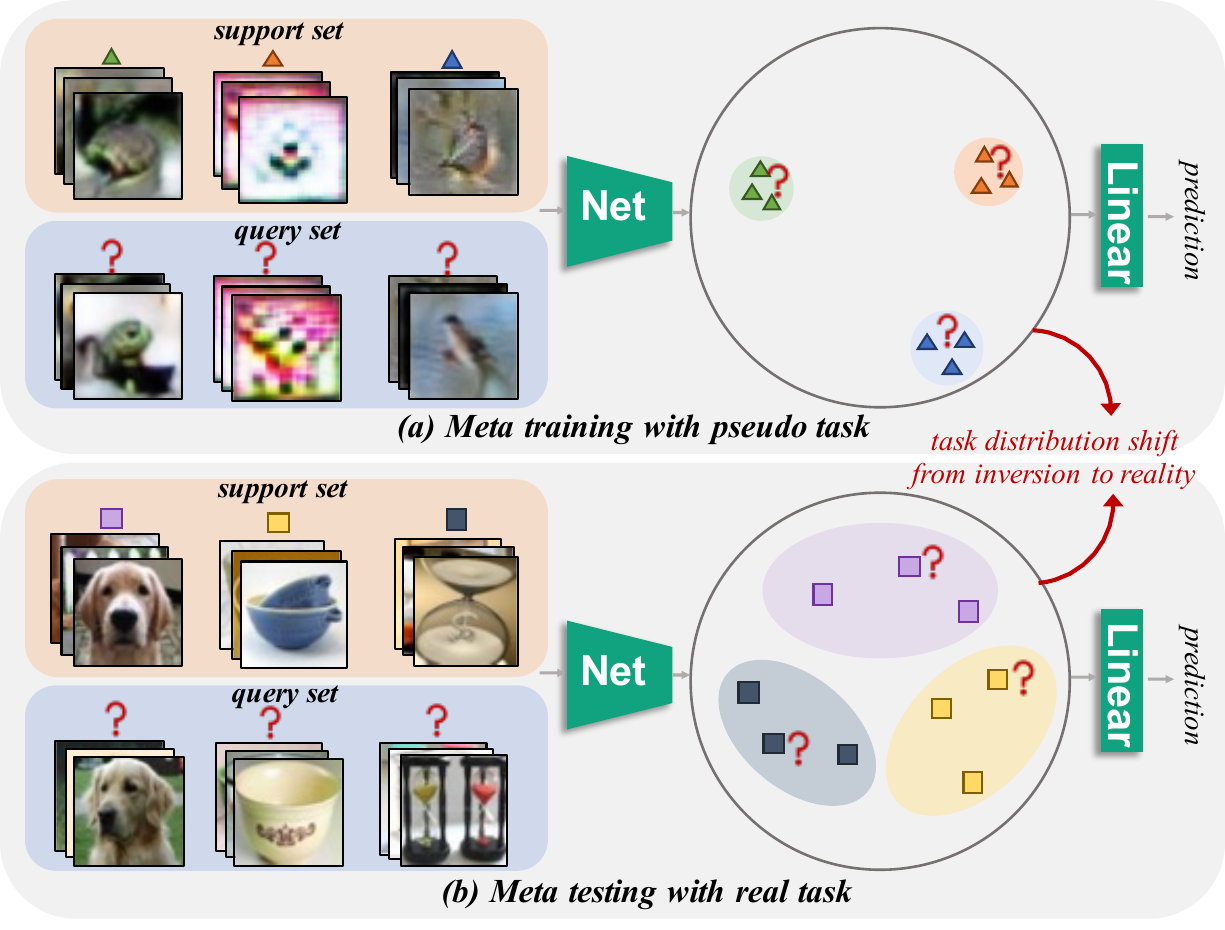}

   \caption{Task-distribution shift between meta training and testing. The pseudo data (as visualized in \cite{fang2021contrastive}) distilled from pre-trained models only contains partial semantic information learned by pre-trained models.}
   \label{fig:shift}
   \vspace{-0.5cm}
\end{figure}

Existing data-free meta-learning methods address the problem in the parameter space. Wang \etal \cite{wang2022metalearning} propose to meta-learn a black-box neural network by predicting the model parameters given the task embedding without data, which can be generalized to unseen tasks. The predicted model parameters with the average task embedding as input are served as the meta initialization for meta testing. However, this method has several drawbacks. First, they only merge the model in parameter space and ignore the underlying data knowledge that could be distilled from the pre-trained models. Second, their method can only be applied to small-scale pre-trained models since they use a neural network to predict the model parameters. Furthermore, their application scenarios are restricted to the case where all pre-trained models have the same architecture, limiting the real-world applicable scenarios.

In this work, we try to address all the above issues simultaneously in a unified framework, named \textbf{PURER} (see \cref{fig:pipeline}), which contains: (1) e\textbf{P}isode c\textbf{U}rriculum inve\textbf{R}sion (ECI) during data-free meta training; and (2) inv\textbf{E}rsion calib\textbf{R}ation following inner loop (ICFIL) during meta testing, thus significantly expanding the application scenarios of DFML. During meta training, we propose ECI to perform pseudo episode training for learning to adapt fast to new unseen tasks. We progressively synthesize a sequence of pseudo episodes (tasks) by distilling the training data from each pre-trained model. ECI adaptively increases the difficulty level of pseudo episode according to the real-time feedback of the meta model. Specifically, we first introduce a small learnable dataset, named \textit{dynamic dataset}. We initialize the dynamic dataset as Gaussian noise and progressively update it to better quality via one-step gradient descent for every iteration. For each episode, we construct a pseudo task by first sampling a subset of labels, and then sampling corresponding pseudo support data and query data. To improve the efficiency of pseudo episode training (see \cref{fig:curriculum}), we introduce the \textit{curriculum mechanism} to synthesize episodes with an increasing level of difficulty. We steer the dynamic dataset towards appropriate difficulty so that only tasks not learned yet are considered at each iteration, which avoids repeatedly synthesizing the tasks already learned well. We design a \textit{Gradient Switch} controlled by the real-time feedback from current meta model, to synthesize harder tasks only when the meta model has learned well on most tasks sampled from current dynamic dataset (see \cref{fig:pipeline}).  Finally, we formulate the optimization process of meta training with ECI as an adversarial form in an end-to-end manner. We further propose a simple plug-and-play supplement---ICFIL---only used during meta testing to narrow the gap between meta training and meta testing task distribution (see \cref{fig:shift}). Overall, our proposed PURER can solve the 
DFML problem using the underlying data knowledge regardless of the dataset, scale and architecture of pre-trained models.

Our method is architecture, dataset and model-scale agnostic, thus substantially expanding the application scope of DFML in real-world applications. We perform extensive experiments in various scenarios, including \textbf{(i)} \textbf{SS}: DFML with \textbf{S}ame dataset and \textbf{S}ame model architecture; \textbf{(ii)} \textbf{SH}: DFML with \textbf{S}ame dataset and \textbf{H}eterogeneous model architectures; \textbf{(iii)} \textbf{MH}: DFML with \textbf{M}ultiple datasets and \textbf{H}eterogeneous model architectures. For benchmarks of SS, SH and MH on CIFAR-FS and MiniImageNet, our method achieves significant performance gains in the range of $6.92\%$ to $17.62\%$, $6.31\%$ to $27.49\%$ and $7.39\%$ to $11.76\%$, respectively.

We summarize the main contributions as three-fold:
\begin{itemize}
    \item We propose a new orthogonal perspective with respect to existing works to solve the data-free meta-learning problem by exploring the underlying data knowledge. Furthermore, our framework is architecture, dataset and model-scale agnostic, \ie, it can be easily applied to various real-world scenarios.
    \item We propose a united framework, PURER, consisting of: (i) ECI to perform pseudo episode training with an increasing level of difficulty during meta training; (ii) ICFIL to narrow the gap between meta training and testing task distribution during meta testing.
    \item Our method achieves superior performance and outperforms the SOTA baselines by a large margin on various benchmarks of SS, SH and MH, which shows the effectiveness of our method.
\end{itemize}
\section{Related Works}
\label{relatedWorks}

\noindent
\textbf{Meta-Learning.}
Meta-Learning \cite{schmidhuber1987evolutionary,naik1992meta,bengio2013optimization} aims to learn general prior knowledge from a large collection of tasks such that the prior can be adapted fast to new unseen tasks. Most existing works \cite{finn2017model, vinyals2016matching,wang2021meta,finn2019online,finn2018probabilistic,li2018learning,yao2021meta,wang2022meta,harrison2020continuous,hou2019cross,zhang2018metagan,zhang2019variational,li2019few,ye2020few,li2020boosting,yang2021free,simon2022meta,yin2019meta,liu2019learning,yoon2018bayesian,hsu2018unsupervised,khodak2019adaptive} assume that the data associated with each task is available during meta training. Recently, data-free meta-learning \cite{wang2022metalearning} problem has attracted researchers' attention. Wang \etal \cite{wang2022metalearning} propose to predict the meta initialization through a meta-trained black-box neural network. However, this method has several drawbacks. First, they focus on parameter space and ignore the underlying data knowledge that could be distilled from the pre-trained models. Second, their method can only be applied for small-scale pre-trained models since they use a neural network to predict the meta initialization. Furthermore, their application scenarios are only restricted to the case where all the pre-trained models have the same architecture, which reduces the applicable scenarios in real applications.  

\noindent
\textbf{Model Inversion.}
Model inversion (MI) \cite{fredrikson2015model,wu2016methodology,zhang2020secret} aims to reconstruct training data from a pre-trained model. Existing works \cite{fredrikson2015model,wu2016methodology,zhang2020secret,deng2021graph,lopes2017data,chawla2021data,zhu2021data,liu2021data,zhang2022fine,fang2021contrastive} synthesize images from pre-trained models via gradient descent. Those works solely synthesize pseudo images without using external feedback, which causes the generated data to be suboptimal or unuseful for meta training. In contrast, our method uses the feedback from meta-learning model to adaptively adjust the synthesized images in an end-to-end way.

\noindent
\textbf{Curriculum Learning.}
Curriculum learning (CL) \cite{bengio2009curriculum,asurvey,soviany2022curriculum} aims to train a model in a meaningful order during loading training data, from easy to hard. 
Kumar \etal \cite{kumar2010self} propose self-paced learning (SPL) to provide automatic curriculum with a loss-based difficulty measure. Recently, Zhang \etal \cite{zhang2021curriculum} propose a curriculum-based meta-learning method by forming harder tasks from each cluster, which merely applies to data-based meta-learning because it needs K-means clustering \cite{lloyd1982least} before training. While in this work, we adaptively synthesize the curriculum on the fly under data-free setting to improve the efficiency of pseudo episode training. 
\section{Problem Setup}
\label{sec:problem-setup}

In this section, we first clarify the definition of Data-free Meta-Learning (DFML) problem. We then discuss the meta testing procedure for DFML problem.

\subsection{Data-free Meta Learning Setup}

We are given a collection of pre-trained models. Each pre-trained model is trained to solve a specific task without accessing their training data. The goal of 
DFML is to learn general prior knowledge (\eg, sensitive  initialization) which can be adapted fast to new unseen tasks.
Here, we emphasize that the pre-trained models may have different architectures. Our method can work with arbitrary architectures of pre-trained models.

\subsection{Meta Testing}

During meta testing, several $N$-way $K$-shot tasks arrive together, which are called the target tasks. The classes appearing in the target tasks have never been seen during both pre-training and meta training. Each task contains a support set with $N$ classes and $K$ instances per class. 
The support set is used for adapting the meta initialization to the specific task. The query set is what the model actually needs to predict. 
Our goal is to adapt the model to the support set so that it can perform well on the query set. The final accuracy is measured by the average accuracy for these target tasks.

\section{Methodology}
\label{sec:methodology}

\begin{figure*}[t]
  \centering
    \includegraphics[width=0.7\linewidth]{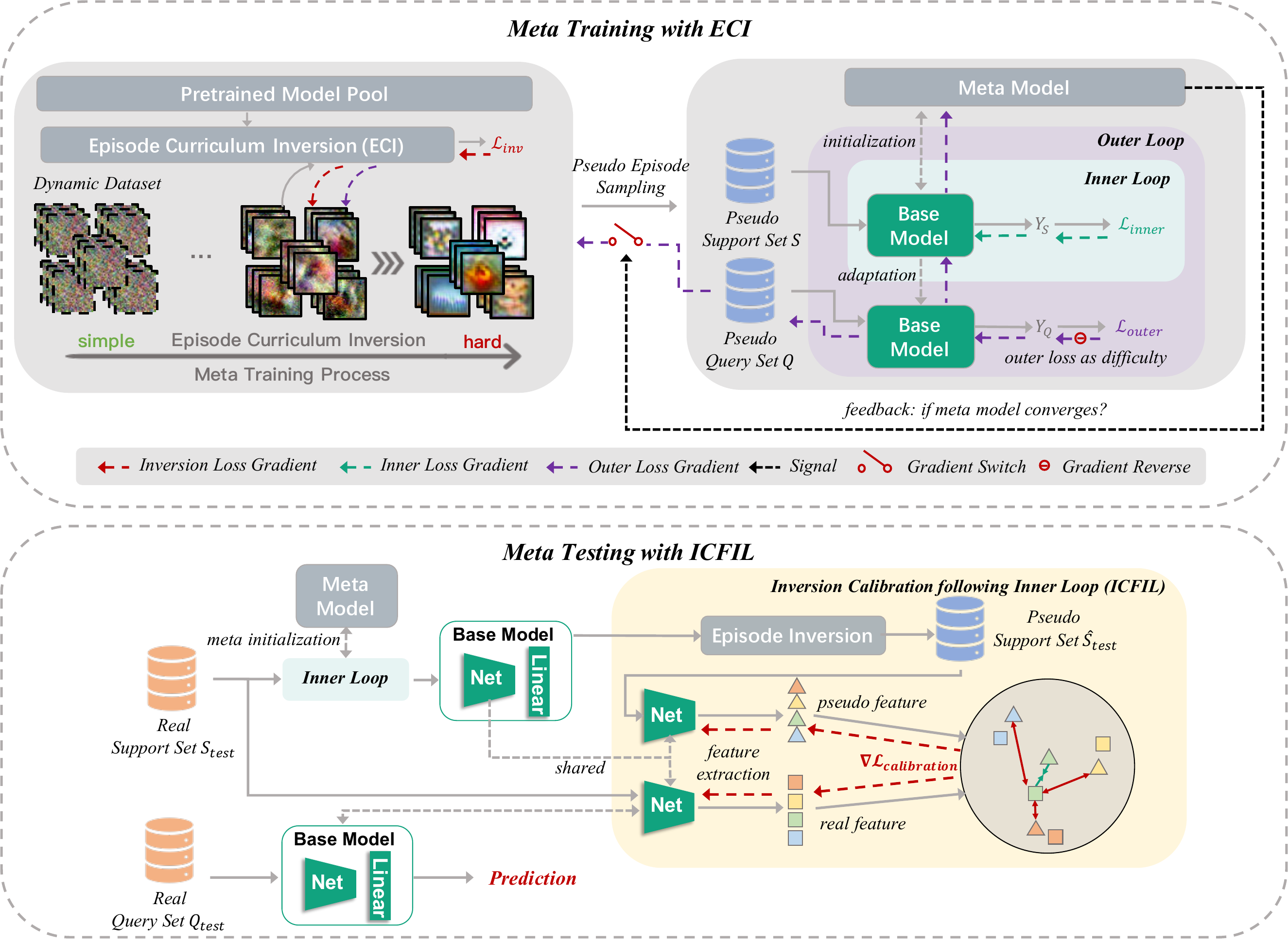}

   \caption{The overall pipeline of our proposed PURER consisting of ECI and ICFIL. For each episode during meta training, a pseudo episode is sampled from the dynamic dataset. The split pseudo support set and query set are used for the inner loop and outer loop of meta-learning. The real-time feedback of meta model controls the Gradient Switch. When the feedback is positive, the dynamic dataset is updated to synthesize harder tasks with larger outer loss for the next iteration by minimizing the reversed outer loss through gradient descent. During meta testing, the adapted base model after inner loop is calibrated via ICFIL. For brevity, we leave out the calibration for linear classifier head.}
   \label{fig:pipeline}
   \vspace{-0.4cm}
\end{figure*}

In this section, we propose a unified framework PURER to solve DFML as illustrated in \cref{fig:pipeline}. In Section \ref{subsec:episodeCurriculumInversion}, we propose ECI to perform pseudo episode training by progressively synthesizing a sequence of pseudo episodes with an increasing level of difficulty during meta training. In Section \ref{subsec:IFCIL}, we further propose ICFIL to eliminate the task-distribution shift issue during meta testing.

\subsection{Preliminary: Episode Training}
\label{subsec:startFromEpisodeTraining}

Our work aims to synthesize a sequence of pseudo episodes to perform pseudo episode training. Thus, we would like to introduce episode training briefly. 
In the case of MAML\cite{finn2017model}, each episode involves:

\begin{itemize}
    \item \textit{Outer Loop}: Update the \textit{meta model} with the goal of improving the performance of \textit{base model} on query set. The meta model works across episodes and learns good generalization from many episodes.
    
    \item \textit{Inner Loop}: Perform \textit{fast adaptation}. The \textit{base model} takes the \textit{meta model} as initialization and performs few steps of gradient descent over the support set. The \textit{base model} works at the level of individual task.
\end{itemize}

Considering a per datum classification loss $l: \mathcal{X} \times \mathcal{Y} \times \Theta \rightarrow \mathbb{R}_{+}$, the \textit{empirical loss} over a finite dataset $\mathcal{B}$ is defined as $\mathcal{L}(\mathcal{B};\boldsymbol{\theta})\triangleq\frac{1}{\left| \mathcal{B} \right|} \sum_{(\boldsymbol{x},y) \in \mathcal{B}} l(\boldsymbol{x},y; \boldsymbol{\theta})$. Thus, for a given task $\mathcal{T}=\{\mathcal{S},\mathcal{Q}\}$, it becomes $\mathcal{L}(\mathcal{S}; \boldsymbol{\theta})$ and $\mathcal{L}(\mathcal{Q}; \boldsymbol{\theta})$. 

 The bi-level optimization can be formulated as follows 

\begin{subequations}
\small
\begin{align}
    \min_{\boldsymbol{\theta}}&\ \mathcal{L}_{outer}(\mathcal{T};\boldsymbol{\theta})\triangleq\mathcal{L}\left(\mathcal{Q}; \boldsymbol{\theta}^{*}\right),\\
     {\rm s.t.}\ 
    \boldsymbol{\theta}^{*} 
    &=\boldsymbol{\theta}-\alpha_{inner}\nabla_{\boldsymbol{\theta}}\mathcal{L}_{inner}(\mathcal{T};\boldsymbol{\theta}) \nonumber\\
    &\triangleq\boldsymbol{\boldsymbol{\theta}}-\alpha_{inner}\nabla_{\boldsymbol{\theta}}\mathcal{L}(\mathcal{S};\boldsymbol{\boldsymbol{\theta}}).\label{subeq:inner}
\end{align}
\label{eq:lossMeta}
\end{subequations}

where $\alpha_{inner}$ is the step size in the inner loop.

\subsection{Episode Curriculum Inversion (ECI)}
\label{subsec:episodeCurriculumInversion}

We propose an ECI component to perform pseudo episode training for learning to adapt fast to new unseen tasks (see \cref{fig:pipeline}). We synthesize a sequence of pseudo episodes by distilling the training data from each pre-trained model. ECI adaptively increases the difficulty level of pseudo episodes according to the real-time feedback of the meta model.

\noindent
\textbf{Episode Inversion.}\ We first propose the basic Episode Inversion (EI) to synthesize the pseudo task for each episode. At the beginning of meta training, we first introduce a small dynamic dataset $\mathcal{D}$ initialized by Gaussian noise with mean $\mu=0$ and standard deviation $\sigma=1$. For each class, it only contains $K+M$ instances, consistent with $N$-way $K$-shot target task configuration with $M$ instances each class in $\mathcal{Q}$. At each iteration, $\mathcal{D}$ is dynamically updated to better quality by taking one-step gradient descent to minimize
\begin{equation}
\small
\mathcal{L}_{inv}(\mathcal{D})=\sum_{(\boldsymbol{\hat{x}},y) \in \mathcal{D}}l(\boldsymbol{\hat{x}},y;\boldsymbol{\psi})+\mathcal{R}_{prior}(\boldsymbol{\hat{x}})+\mathcal{R}_{feature}(\boldsymbol{\hat{x}}).
\label{eq:lossData}
\end{equation}
$\boldsymbol{\psi}$ is the parameter of the corresponding pre-trained model. For each pair $(\boldsymbol{\hat{x}},y)$, $\boldsymbol{\hat{x}}$ is the pseudo image, and $y$ is the manually assigned label to which we desire $\boldsymbol{\hat{x}}$ to belong. $l(\cdot)$ is the per datum classification loss (\eg, cross-entropy loss) mentioned in \cref{subsec:startFromEpisodeTraining}. $\mathcal{R}_{prior}(\cdot)$ is borrowed from DeepDream \cite{deepdream} to steer $\hat{x}$ away from unreal images:

\begin{equation}
\small
\mathcal{R}_{prior}(\boldsymbol{\hat{x}})=\alpha_{TV}\mathcal{R}_{TV}(\boldsymbol{\hat{x}})+\alpha_{l_{2}}\mathcal{R}_{l_{2}}(\boldsymbol{\hat{x}}),
\label{eq:prior}
\end{equation}
where $\mathcal{R}_{TV}$ and $\mathcal{R}_{l_{2}}$ are the total variance and $l_2$ norm of $\boldsymbol{\hat{x}}$, with scaling factor $\alpha_{TV}$ and $\alpha_{l_{2}}$.
$\mathcal{R}_{feature}(\cdot)$ is a feature distribution regularization term used in DeepInversion \cite{yin2020dreaming} to minimize the distance between feature maps of pseudo images and original training images in each convolutional layer:
\begin{equation}
\small
\begin{aligned}
\mathcal{R}_{feature}(\boldsymbol{\hat{x}})=&\sum_{l}\Vert\mu_{l}(\boldsymbol{\hat{x}})-{\rm BN}_{l}\rm{(running\_mean)}\Vert+\\
&\sum_{l}\Vert\sigma_l^2(\boldsymbol{\hat{x}})-{\rm BN}_{l}{\rm (running\_variance)}\Vert.
\end{aligned}
\label{eq:feature}
\end{equation}
We feed a batch of pseudo images into the given pre-trained model to obtain feature maps in each convolutional layer. $\mu_{l}(\boldsymbol{\hat{x}})$ and $\sigma_l^2(\boldsymbol{\hat{x}})$ are the batch-wise mean and variance of those feature maps in the $l^{th}$ convolutional layer. ${\rm BN}_{l}\rm{(running\_mean)}$ and ${\rm BN}_{l}{\rm (running\_variance)}$ are the running average mean and variance stored in the $l^{th}$ BatchNorm layer of pre-trained model, which is calculated during the pre-training period using original training images and can be obtained without access to original training images.

For each episode, we construct a pseudo task by first sampling a subset of labels, and then sampling the pseudo support data and query data from $\mathcal{D}$.

\noindent
\textbf{Curriculum Mechanism.}\ Directly synthesizing the pseudo images without any feedback may repeatedly generate not optimal or not useful data for meta training (see \cref{fig:curriculum}). For example, we may repeatedly synthesize the tasks already learned well, thus increasing the unnecessary inversion cost. To find an effective episode inversion path, we propose to perform EI with curriculum mechanism, \ie ECI. We aim to steer the dynamic dataset $\mathcal{D}$ towards appropriate difficulty so that only tasks not learned yet are considered at each iteration, which avoids repeatedly synthesizing the tasks already learned well. We characterize the difficulty of $\mathcal{D}$ for current meta model (parameterized by $\boldsymbol{\theta}$) as the expected outer loss over candidate tasks, \ie $\underset{\mathcal{T} \in \mathcal{D}}{\mathbb{E}}\left[\mathcal{L}_{outer}(\mathcal{T};\boldsymbol{\theta})\right]$. Specifically, a large outer loss presents a hard task; a task which has been learned well by current meta model, however, leads to a small outer loss. 
Thus, as illustrated in \cref{fig:pipeline}, we design a loss-based criteria to judge whether the meta model has learned well on tasks sampled from current $\mathcal{D}$. Specifically, at each iteration, for a batch of episodes $\{\mathcal{T}_{i}\}$, if the sum of the outer loss (training accuracy) over $\{\mathcal{T}_{i}\}$ has not decreased (increased) for 6 consecutive iterations, the meta model will send a positive feedback, indicating that harder tasks are needed for the next iteration; else negative. To synthesize harder tasks only when the feedback is positive, we design the \textit{Gradient Switch} controlled by the real-time feedback. Specifically, if the meta model has converged, the positive feedback will turn on the Gradient Switch so that the reversed gradient of outer loss can flow backwards to $\mathcal{D}$ (see \cref{fig:pipeline}). In this way, we will update $\mathcal{D}$ to maximize the outer loss through gradient descent, which means we can obtain harder tasks with a larger outer loss for the next iteration.
The Gradient Switch controlled by feedback $\Omega$ works as an indicator function, \ie, 
\begin{equation}
\small
\mathbb{I}(\Omega)=
\left\{
\begin{aligned}
    &1,\  if\ \Omega\ is\ positive; \\
    &0,\  if\ \Omega\ is\ negative.
\end{aligned}
\right.
\label{eq:switch}
\end{equation}
In this way, ECI finds an effective episode inversion path by ignoring the tasks already learned well, thus improving the efficiency of pseudo episode training.

\begin{algorithm}[t]
\small
\DontPrintSemicolon
\SetKwInOut{Input}{Input}\SetKwInOut{Output}{Output}\SetKwInOut{Require}{Require}
\textbf{REQUIRE} Meta-model parameters $\boldsymbol{\theta}$;  dynamic dataset $\mathcal{D}$; step size $\alpha_{inner}$, $\alpha_{outer}$ and $\beta$; scaling factor $\lambda$.\;
Initialize $\mathcal{D}$\ and $\boldsymbol{\theta}$\;
\While(){not done}{
\tcp{Episode Curriculum Inversion}
Evaluate $\-\mathcal{L}_{inv}(\mathcal{D})$ \wrt \cref{eq:lossData}\;
\If {feedback $\Omega$\ is\ Positive}{
Randomly sample a task $\mathcal{T}$\ from $\mathcal{D}$\;
Evaluate $\mathcal{L}_{outer}(\mathcal{T}; \boldsymbol{\theta})$ \wrt \cref{eq:lossMeta}\;
Update $\mathcal{D} \leftarrow \mathcal{D} - \beta*\nabla_{\mathcal{D}}\left(\mathcal{L}_{inv}(\mathcal{D})-\lambda*\mathcal{L}_{outer}(\mathcal{T}; \boldsymbol{\theta})\right)$
}
\Else{Update $\mathcal{D} \leftarrow \mathcal{D} - \beta*\nabla_{\mathcal{D}}\left(\mathcal{L}_{inv}(\mathcal{D})\right)$}
\tcp{Meta Training}
Randomly generate a batch of tasks $\mathcal{T}_i$ from $\mathcal{D}$\;
\For{$\mathbf{all}$ $\mathcal{T}_i$}{
Evaluate $\mathcal{L}_{outer}(\mathcal{T}_i; \boldsymbol{\theta})$ \wrt \cref{eq:lossMeta}\;
}
Update $\boldsymbol{\theta} \leftarrow \boldsymbol{\theta} - \alpha_{outer}*\nabla_{\boldsymbol{\theta}}\sum_{\mathcal{T}_i}\mathcal{L}_{outer}(\mathcal{T}_i; \boldsymbol{\theta})$\;
\tcp{Update Feedback}
Check \textit{feedback} $\Omega$\;
}
\caption{Data-free Meta Training with ECI\label{IR}}
\label{alg:eci}
\end{algorithm}

 \noindent
\textbf{Adversarial Optimization.}\ 
To adaptively synthesize episodes from easy to hard along with the meta training process,
we perform pseudo episode training by adversarially updating the meta model (parameterized by $\boldsymbol{\theta}$) and dynamic dataset $\mathcal{D}$ in an end-to-end manner. We search the optimal meta model parameters $\boldsymbol{\theta}$ by minimizing the expected outer loss  over candidate tasks from $\mathcal{D}$ , \ie, $\underset{\mathcal{T} \in \mathcal{D}}{\mathbb{E}}[\mathcal{L}_{outer}(\mathcal{T}; \boldsymbol{\theta})]$, while 
 we update the dynamic dataset $\mathcal{D}$ to synthesize harder tasks with higher quality by minimizing reversed outer loss $-\underset{\mathcal{T} \in \mathcal{D}}{\mathbb{E}}[\mathcal{L}_{outer}(\mathcal{T}; \boldsymbol{\theta})]$ (when feedback $\Omega$ is positive) and inversion loss $\mathcal{L}_{inv}(\mathcal{D})$. Thus, we formulate the overall optimization process as an adversarial form:
\begin{equation}
\small
\begin{aligned}
\min_{\boldsymbol{\theta}}\max_{\mathcal{D}}\underset{\mathcal{T} \in \mathcal{D}}{\mathbb{E}}[-\mathcal{L}_{inv}(\mathcal{D})
+\mathbb{I}(\Omega)*\mathcal{L}_{outer}(\mathcal{T};\boldsymbol{\theta})].
\end{aligned}
\label{eq:lossAll}
\end{equation}
To the end, we summarize the detailed procedure of ECI for data-free meta training in \cref{alg:eci}.

\begin{algorithm}[t]
\small
\DontPrintSemicolon
\SetKwInOut{Input}{Input}\SetKwInOut{Output}{Output}\SetKwInOut{Require}{Require}
\textbf{REQUIRE} The meta testing task $\mathcal{T}_{test}=\{\mathcal{S}_{test},\mathcal{Q}_{test}\}$; meta initialization $\boldsymbol{\theta}$.\;
\tcp{Inner Loop}
Obtain $\{\boldsymbol{\varphi}_{test}^{*},\mathbf{W}_{test}^{*}\}$ via fast adaptation over $\mathcal{S}_{test}$\;
Obtain pseudo support set $\hat{\mathcal{S}}_{test}$ \wrt \cref{eq:lossData}\;
\tcp{Calibration}
Calibrate $\boldsymbol{\varphi}_{test}^{*}$ by minimizing \cref{eq:contrast}\;
Train a new linear classifier head\;
\tcp{Prediction}
Make predictions on the query set $\mathcal{Q}_{test}$
\caption{Meta Testing with ICFIL}
\label{alg:icfil}
\end{algorithm}

\subsection{Inversion Calibration following Inner Loop} \label{subsec:IFCIL}
Pseudo images synthesized via ECI can not cover all semantic information of real images, thus leading to a gap between meta training and testing task distribution (see \cref{fig:shift}). In other words, the task-distribution shift issue in the data-free setting is much more significant than that in the setting of data-based meta-learning. To address this issue, we propose a simple plug-and-play supplement---ICFIL---only used during meta testing, to further narrow the gap between meta training and testing task distribution.

For an $N$-way target task $\mathcal{T}_{test}=\{\mathcal{S}_{test},\mathcal{Q}_{test}\}$ during meta testing, we are given the meta initialization $\boldsymbol{\theta}$ according to \cref{alg:eci}. The base model initialized by $\boldsymbol{\theta}$ comprises a backbone $\phi:\mathbb{R}^{N_{i n}} \rightarrow \mathbb{R}^{N_f}$ (parameterized by $\boldsymbol{\varphi}$) and a linear classifier head parametrized by the weight matrix $\mathbf{W} \in \mathbb{R}^{N_f \times N}$, where $N_{in}$, $N_f$ are the dimension of the input and embedding, respectively. We first perform fast adaptation over $\mathcal{S}_{test}$. The adapted parameters are $\boldsymbol{\theta}_{test}^{*}=\{\boldsymbol{\varphi}_{test}^{*},\mathbf{W}_{test}^{*}\}$. Note that this is precisely identical to the standard meta testing procedure for regular meta-learning. The ICFIL does not assume access to query set $\mathcal{Q}_{test}$ or any additional data and it can be viewed as a simple plug-and-play supplement for the standard meta testing period. We use ICFIL to calibrate the adapted base model before it predicts on the query set $\mathcal{Q}_{test}$. As illustrated in \cref{fig:pipeline}, we first synthesize a batch of pseudo images $\hat{\mathcal{S}}_{test}$ from the adapted base model according to \cref{eq:lossData}. Inspired by the supervised contrastive learning \cite{khosla2020supervised}, for a given image $\boldsymbol{x}$ in $\mathcal{S}_{test}$, we define all pseudo images $\boldsymbol{\hat{x}}$ with the same label as positive samples $\boldsymbol{\hat{x}}^{+}$ while the others as negative samples $\boldsymbol{\hat{x}}^{-}$. The core idea behind ICFIL is to force the base model to focus on the overlapping information in both pseudo and real images, and ignore the others. 
The calibration loss is formulated as:
\begin{equation}
\small
\!\!\! \mathcal{L}_{calibration}(\phi)\!=\!-\!\sum_{\boldsymbol{x} \in \mathcal{S}_{test}}\sum_{\boldsymbol{\hat{x}}^{+}}\log\frac{\exp\left[\phi(\boldsymbol{x})^{T}\phi(\boldsymbol{\hat{x}}^{+})\//\tau\right]}{\sum_{\boldsymbol{\hat{x}}}\exp\left[\phi(\boldsymbol{x})^{T}\phi(\boldsymbol{\hat{x}})\//\tau\right]}.
\label{eq:contrast}
\end{equation}
where temperature parameter $\tau$ is widely used to help discriminate positive and negative samples. We then freeze the backbone and train a new linear classifier head by minimizing a standard classification loss (\eg, cross-entropy loss) over $\mathcal{S}_{test}$. The whole procedure for meta testing with ICFIL is provided in \cref{alg:icfil}.

\section{Experiments}
\label{sec:experiments}

We conduct extensive experiments in various real-world scenarios to demonstrate the effectiveness of our proposed PURER framework.
In \cref{subsec:setup}, we present the basic experiment setup. In \cref{subsec:ss,subsec:sh,subsec:mh}, we provide experiments about DFML in various scenarios, including SS, SH and MH. In \cref{subsec:ablation}, we provide more analysis of PURER.

\subsection{Experiments Setup}
\label{subsec:setup}
\noindent
\textbf{Baselines.}\ We compare PURER with four typical baselines: \textbf{ (i) Random.}\ Randomly initialize the base model before the adaptation of each target task. \textbf{(ii) Average.}\ Average all pre-trained models to initialize the base model before the adaptation of each target task. \textbf{(iii) OTA\cite{singh2020model}.}\ Calculate the weighted average of all pre-trained models via optimal transport as the initialization of the base model before the adaptation of each target task. 
\textbf{(iv) DRO\cite{wang2022metalearning}.}\ Meta-learn a neural network to predict the model parameters given the task embedding. The predicted model parameters with the average task embedding as input are served  as the meta initialization. This is the first work to solve DFML by directly predicting the meta initialization in parameter space.

\noindent
\textbf{Datasets for meta testing.}\ 
CIFAR-FS \cite{bertinetto2018meta} and MiniImageNet \cite{vinyals2016matching} are commonly used in meta-learning, which consist of 100 classes with 600 images per class, respectively. We adopt a standard dataset split as \cite{wang2022metalearning}: 64 classes for meta training, 16 classes for meta validation and 20 classes for meta testing. All splits are non-overlapping. 
Note that we have no access to the meta training data in DFML setting.

\noindent
\textbf{Evaluation metric.}\ 
We evaluate the performance by the average accuracy and standard deviation over 600 unseen target tasks sampled from meta testing dataset.

\noindent
\textbf{Scenarios.}\ (i) \textbf{SS}. All pre-trained models are trained on \textbf{S}ame dataset with \textbf{S}ame architecture. (ii) \textbf{SH}. All pre-trained models are trained on \textbf{S}ame dataset but with \textbf{H}eterogeneous architectures. (iii) \textbf{MH}. All pre-trained models are trained on \textbf{M}ultiple datasets with \textbf{H}eterogeneous architectures. Existing DFML work, \ie, baseline DRO \cite{wang2022meta} and baselines Average and OTA 
merely apply to SS because they require pre-trained models have the same architecture. We argue that SH and MH are more widely applicable scenarios in real-world applications and our PURER is a unified framework for addressing SS, SH and MH simultaneously.

\subsection{Experiments of DFML in SS}
\label{subsec:ss}
\label{subsec:same}
\noindent
\textbf{Overview.}\ We first perform experiments in \textbf{SS} scenario. 
We construct a collection of $N$-way tasks from the meta-training dataset and pre-train the model via standard supervised learning for each task. The collection of pre-trained models is used as the given resource models for DFML.

\noindent
\textbf{Implementation details.}\ We take Conv4 
as the architecture of all pre-trained models and the meta model, the same as regular meta-learning works\cite{finn2017model,chen2020variational,liu2020adaptive,wang2019simpleshot}.  We take one-step gradient descent to perform fast adaptation for both meta training and testing. More implementation details are provided in \cref{app:implementationdetails}.

\begin{table}[htbp]
  \centering
  \small
   \caption{Compare to baselines in SS scenario.} 
   \vspace{-0.2cm}
  \begin{tabular}{clcc}
    \toprule
     \textbf{SS}&\textbf{Method}  & 1-shot &5-shot \\
    \midrule
    \multirow{5}{*}{\shortstack{\textbf{CIFAR-FS}\\\textbf{5-way}}} & Random &21.65  $\pm$ 0.45 & 21.59  $\pm$ 0.45\\
     & Average & 28.12  $\pm$ 0.62& 32.15  $\pm$ 0.64\\
     & OTA & 29.10 $\pm$ 0.65 & 34.33 $\pm$ 0.67\\
     & DRO & 23.92 $\pm$ 0.49 & 24.34 $\pm$ 0.49\\
     
     & Ours & \textbf{38.66 $\pm$ 0.78} & \textbf{51.95 $\pm$ 0.79}\\
    \midrule
    \midrule
    \multirow{5}{*}{\shortstack{\textbf{MiniImageNet}\\\textbf{5-way}}} & Random &22.45  $\pm$ 0.41 & 23.48  $\pm$ 0.45\\
     & Average & 22.87  $\pm$  0.39&26.13  $\pm$  0.43\\
     & OTA & 24.22 $\pm$ 0.53& 27.22 $\pm$ 0.59 \\
     & DRO & 23.96 $\pm$ 0.42 &25.81 $\pm$ 0.41\\

     & Ours & \textbf{31.14 $\pm$ 0.63} & \textbf{40.86 $\pm$ 0.64}\\
     \bottomrule
  \end{tabular}
  \label{tab:dfmeta5way}
\end{table}

\noindent
\textbf{Results.}\ \cref{tab:dfmeta5way} shows the results for 5-way classification on CIFAR-FS and MiniImageNet under the scenario of SS. For CIFAR-FS, our method outperforms the best baseline by $9.56\%$ and $17.62\%$ for 1-shot and 5-shot learning, respectively. For MiniImageNet,  our method outperforms the best baseline by $6.92\%$ and $13.64\%$ for 1-shot and 5-shot learning, respectively. The results show that random initialization is insufficient to perform fast adaptation. Simply averaging and OTA perform worse because: i) they fuse models layer-wise, but each pre-trained model are trained to solve different tasks, thus lacking the precise correspondence in parameter space; ii) they lack meta-learning objectives so that the fused model can not generalize well to new unseen tasks. DRO also performs not better due to the small number of pre-trained models. DRO trains a neural network to predict the meta initialization, thus requiring a relatively large number of pre-trained models as training resources. Our proposed method performs best because we leverage the fruitful underlying data knowledge contained in each pre-trained model instead of focusing on parameter space. We perform pseudo episode training with meta-learning objectives, thus learning to adapt fast to new unseen tasks with few labeled data.

\subsection{Experiments of DFML in SH}
\label{subsec:sh}
\noindent
\textbf{Overview.} To verify the broad applicability of our proposed method, we perform experiments under a more realistic and challenging scenario, \textbf{SH}. For each task, we pre-train the model with a randomly selected architecture.

\noindent
\textbf{Implementation details.}\ For each task, we pre-train the model with an architecture randomly selected from Conv4, ResNet-10 and ResNet-18. Compared to Conv4, ResNet-10 and ResNet-18 are larger-scale neural networks. We take Conv4 as the meta model architecture. More implementation details are provided in \cref{app:implementationdetails}.

\noindent
\textbf{Results.}\ \cref{tab:dfmetamix} shows the results for 5-way classification on CIFAR-FS and MiniImageNet under the scenario of SH. For CIFAR-FS, our proposed method outperforms the baseline by $17.50\%$ and $27.49\%$ for 1-shot and 5-shot learning, respectively. For MiniImageNet, our method outperforms the baseline by $6.31\%$ and $11.71\%$ for 1-shot and 5-shot learning, respectively. Ours can apply to this real-world scenario and perform the best because we leverage the underlying data knowledge regardless of the scale and architecture of pre-trained models.

\begin{table}[htbp]
  \centering
  \small
  \caption{Compare to baselines in SH scenario. Pre-trained models are trained with heterogeneous architectures (Conv4, ResNet-10 and ResNet-18).}
  \label{tab:dfmetamix}
    \vspace{-0.2cm}
  \begin{tabular}{clcc}
    \toprule
     \textbf{SH}&\textbf{Method}  & 1-shot & 5-shot \\
    \midrule
    \multirow{2}{*}{\shortstack{\textbf{CIFAR-FS}\\\textbf{5-way}}} & Random &21.65  $\pm$ 0.45& 21.59  $\pm$ 0.45  \\
     
     & Ours &\textbf{39.15 $\pm$ 0.70} & \textbf{49.08 $\pm$ 0.74} \\
    \midrule
    \midrule
    \multirow{2}{*}{\shortstack{\textbf{MiniImageNet}\\\textbf{5-way}}} & Random & 22.45  $\pm$  0.41&23.48  $\pm$  0.45 \\
     & Ours & \textbf{28.76 $\pm$ 0.60} & \textbf{35.19 $\pm$ 0.64}\\
     \bottomrule
  \end{tabular}
\end{table}

\subsection{Experiments of DFML in MH}
\label{subsec:mh}
\noindent
\textbf{Overview.}\ We further perform experiments in \textbf{MH} scenario, which provides a more challenging test.  We construct a collection of tasks from different meta training datasets. For each task, we pre-train the model with a randomly selected architecture. In this way, we obtain a collection of pre-trained models trained on various datasets with different architectures.

\noindent
\textbf{Implementation details.}\ During meta training, we construct each task from the meta training dataset randomly selected from CIFAR-FS and MiniImageNet. We pre-train the model with an architecture randomly selected from Conv4, ResNet-10 and ResNet-18. We take Conv4 as the meta model architecture. During meta testing, we report the average accuracy over 600 tasks sampled from both CIFAR-FS and MiniImageNet meta testing datasets. More implementation details are provided in \cref{app:implementationdetails}.

\noindent
\textbf{Results.}\ \cref{tab:dfmetamixdata} shows the results for 5-way classification under the scenario of MH. Our proposed method outperforms the baseline by $7.39\%$ and $11.76\%$ for 5-way 1-shot and 5-shot learning, respectively. Ours can apply to this more challenging scenario because our method is dataset-agnostic and architecture-agnostic. In other word, PURER works well on pre-trained models trained on various datasets with different architectures.

\begin{table}[htbp]
  \centering
  \small
  \caption{Compare to baselines in MH scenario. Pre-trained models are from multiple datasets (CIFAR-100 and MiniImageNet)  with heterogeneous architectures (Conv4, ResNet-10 and ResNet-18).}
  \label{tab:dfmetamixdata}
    \vspace{-0.2cm}
  \begin{tabular}{clcc}
    \toprule
     \textbf{MH}&\textbf{Method}  &  1-shot & 5-shot \\
    \midrule
    \multirow{2}{*}{\shortstack{\textbf{5-way}}} & Random &21.11  $\pm$ 0.41& 21.34  $\pm$ 0.40  \\
     
     & Ours &\textbf{28.50 $\pm$ 0.63} & \textbf{33.10 $\pm$ 0.69} \\
     \bottomrule
  \end{tabular}
\end{table}

\subsection{Ablation Study}
\label{subsec:ablation}

\noindent
\textbf{Effectiveness of each component in PUERE.}\  \cref{tab:ablation} shows the results of ablation studies in SS scenario. We first introduce a vanilla baseline (EI) performing pseudo episode training without curriculum mechanism and meta testing without ICFIL. EI still outperforms the best baselines in \cref{tab:dfmeta5way} by $3.8\%$ and $6.55\%$ for 1-shot and 5-shot learning, respectively, which verifies the effectiveness of the basic idea of DFML leveraging the underlying data knowledge. To further evaluate the effectiveness of curriculum mechanism, we append curriculum to EI, \ie, ECI. We can observe that with curriculum mechanism, the performance can be improved by $1.09\%$ and $2.93\%$ for 1-shot and 5-shot learning, respectively. Similarly, we adopt ICFIL during meta testing for EI, \ie, EI + ICFIL, and the performance can be improved by $1.84\%$ and $2.89\%$ for 1-shot and 5-shot learning, respectively. By introducing both curriculum mechanism for episode inversion and ICFIL for meta testing, we achieve the best performance with a boosting gain of $3.12\%$ and $7.09\%$ for 1-shot and 5-shot learning, respectively, which demonstrates the effectiveness of the joint schema.

\begin{table}[h]
  \centering
  \small
  \caption{Ablation studies on MiniImageNet in SS scenario.}
  \scalebox{0.7}{
  \begin{tabular}{ccccc}
    \toprule
    \small
    \multirow{2}{*}{\textbf{SS}} & \multicolumn{2}{c}{Component} & \multicolumn{2}{c}{Accuracy} \\
    \cmidrule(r){2-3}
    \cmidrule(r){4-5}
    &\textbf{Curriculum}&\textbf{ICFIL}&5-way 1-shot&5-way 5-shot\\
    \midrule
    \textbf{EI}&&&28.02 $\pm$ 0.58&33.77 $\pm$ 0.63\\
    \midrule
    &\checkmark&& 29.11$\pm$ 0.59 &36.70 $\pm$ 0.64 \\
    &&\checkmark&29.86 $\pm$ 0.66&36.66 $\pm$ 0.66\\
    \midrule
    \textbf{Ours}&\checkmark&\checkmark&\textbf{31.14$\pm$ 0.63}&\textbf{40.86 $\pm$ 0.64}\\
     \bottomrule
  \end{tabular}}
  
  \label{tab:ablation}
\end{table}

\begin{table}[tbp]
  \centering
  \small
  \caption{Effect of the number of training class for each task in SS.}
    \label{tab:dfmeta10way}
      \vspace{-0.2cm}
      \scalebox{0.8}{
  \begin{tabular}{clcc}
    \toprule

     \textbf{SS}&\textbf{Method}  &  1-shot &  5-shot \\
    \midrule
    \multirow{5}{*}{\shortstack{\textbf{CIFAR-FS}\\\textbf{10-way}}} & Random &  10.26 $\pm$ 0.22 & 10.32 $\pm$ 0.22 \\
     & Average  & 12.23 $\pm$ 0.30 &14.86 $\pm$ 0.30 \\
     & OTA & 13.40 $\pm$ 0.32 &15.17 $\pm$ 0.36\\

     & DRO & 11.29 $\pm$ 0.25 & 11.56 $\pm$ 0.26 \\

     & Ours & \textbf{22.88 $\pm$ 0.41} & \textbf{36.19 $\pm$ 0.45}\\

    \midrule
    \midrule
    \multirow{5}{*}{\shortstack{\textbf{MiniImageNet}\\\textbf{10-way}}} & Random & 
 11.20 $\pm$ 0.23 & 11.34 $\pm$ 0.24  \\
     & Average & 12.83 $\pm$ 0.27 & 13.94 $\pm$ 0.27 \\
     & OTA & 11.92 $\pm$ 0.25&13.25 $\pm$ 0.29 \\
     & DRO &11.78 $\pm$ 0.20 & 13.08 $\pm$ 0.24 \\
     
     & Ours & \textbf{18.16 $\pm$ 0.37}&\textbf{ 26.92 $\pm$ 0.37}\\

     \bottomrule
  \end{tabular}}
    \vspace{-0.2cm}
\end{table}

\begin{table}[tbp]
  \centering
  \small
  \caption{Compare to MAML in SS scenario. 
  $^{\dag}$: use real meta-training dataset of equal size to the dynamic dataset used in DFML. Grey: data-based meta-learning method.}
  \label{tab:maml}
    \vspace{-0.2cm}
  \scalebox{0.9}{
  \begin{tabular}{clcc}
    \toprule
     \textbf{SS}&\textbf{Method}  & 1-shot & 5-shot \\
    \midrule
    \multirow{2}{*}{\shortstack{\textbf{CIFAR-FS}\\\textbf{5-way}}} & \cellcolor{gray!40}MAML$^{\dag}$ &\cellcolor{gray!40} 34.18 $\pm$ 0.75 &\cellcolor{gray!40} 41.20 $\pm$ 0.71 \\
    &Ours &\textbf{38.66 $\pm$ 0.78} & \textbf{51.95 $\pm$ 0.79}\\
    \midrule
    \midrule
    \multirow{2}{*}{\shortstack{\textbf{CIFAR-FS}\\\textbf{10-way}}} &\cellcolor{gray!40}MAML$^{\dag}$ &\cellcolor{gray!40} 19.05 $\pm$ 0.38 &\cellcolor{gray!40} 21.04 $\pm$ 0.35 \\
    &Ours &\textbf{22.88 $\pm$ 0.41} & \textbf{36.19 $\pm$ 0.45}\\
     \bottomrule
  \end{tabular}}
  \vspace{-0.38cm}
\end{table}

\noindent
\textbf{Effect of the number of training class for each task.} To evaluate the performance difference for the different number of training class for each task, we perform the experiments for 10-way classification problem in SS scenario. \cref{tab:dfmeta10way} shows the results. For CIFAR-FS, our method outperforms the best baseline by $9.48\%$ and $21.02\%$ for 1-shot and 5-shot learning, respectively. For MiniImageNet, our method outperforms the best baseline by $5.33\%$ and $12.98\%$ for 1-shot and 5-shot learning, respectively. Compared to 5-way classification, the accuracy of 10-way classification is lower because it is more challenging. Ours outperforms all baselines in both 5-way and 10-way classification problems.

\noindent
\textbf{Comparison to MAML.}\ 
To evaluate the effectiveness of DFML, we compare our method with regular data-based meta-learning work MAML\cite{finn2017model} under the scenario of SS. The results are shown in \cref{tab:maml}. We introduce MAML$^{\dag}$, which uses the real meta training dataset of equal size to the dynamic dataset used in DFML.  For 5-way classification on CIFAR-FS, our method outperforms MAML$^{\dag}$ by $4.48\%$ and $10.75\%$ for 1-shot and 5-shot learning, respectively. For 10-way classification on CIFAR-FS, our method outperforms MAML$^{\dag}$ by $3.83\%$ and $15.15\%$ for 1-shot and 5-shot learning, respectively. MAML$^{\dag}$ performs worse because  MAML$^{\dag}$ uses a fixed meta training dataset, which only provides a limited number of tasks in a random order. In contrast, PURER can provide a more diverse collection of tasks with an increasing level of difficulty because the dynamic dataset can be adaptively updated according to the real-time feedback of meta-learning model.

\begin{figure}[tbp]
  \centering
    \includegraphics[width=0.99\linewidth]{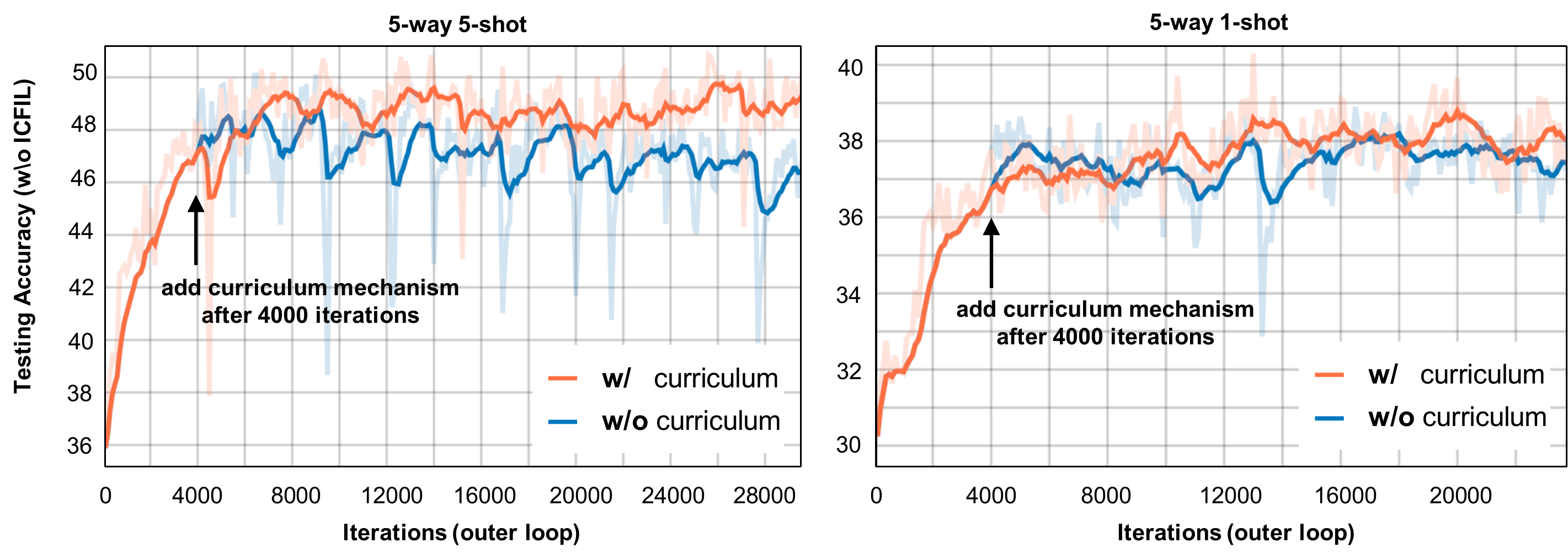}
      \vspace{-0.3cm}
   \caption{Effect of curriculum mechanism on testing performance on CIFAR-FS in SS scenario. 
   Solid curve: smoothed performance curve. Transparent curve: original performance curve.}
   \label{fig:cur}
   \vspace{-0.2cm}
\end{figure}

\noindent
\textbf{Contributions from curriculum mechanism.}\ 
\cref{fig:cur} shows the testing accuracy with and without curriculum mechanism, respectively. After 4000 iterations, we adopt curriculum mechanism to adaptively synthesize harder tasks. Without curriculum mechanism, the performance converges to a lower peak and even begins to decline, which can be explained as overfitting to the tasks of single-level difficulty. In contrast, with curriculum mechanism, the meta model can keep learning from harder tasks and finally achieves a peak with higher testing performance.

\begin{figure}[tbp]
  \centering
    \includegraphics[width=0.9\linewidth]{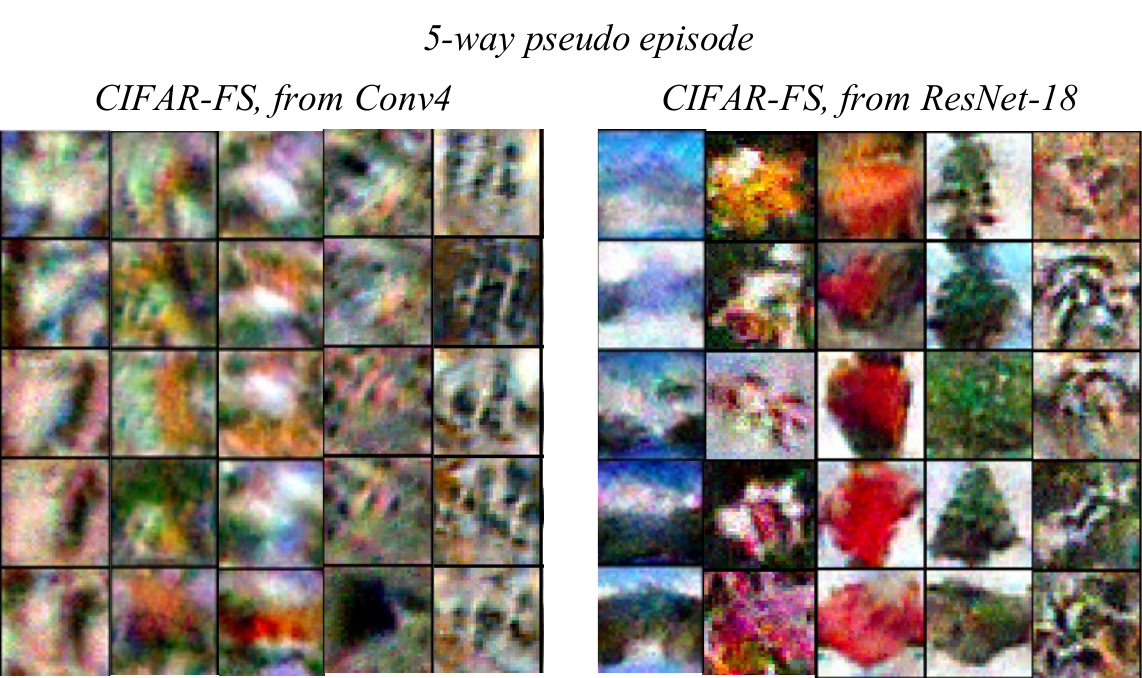}
     \vspace{-0.3cm}
   \caption{  (left) pseudo images synthesized from Conv4. (right) pseudo images synthesized from ResNet-18. Each column corresponds to one class.}
   \label{fig:vis}
   \vspace{-0.4cm}
\end{figure}

\noindent
\textbf{Visualizations of ECI.}\ 
\cref{fig:vis} shows visualizations of 5-way pseudo episodes synthesized by ECI from models pre-trained on CIFAR-FS with architectures of Conv4 and ResNet-18, respectively. We can observe that compared with Conv4, a deeper pre-trained model (ResNet-18) can synthesize images of better quality.

\noindent
\textbf{Hyperparameter sensitivity.}\ 
Results in \cref{app:moreresults} show that the performance of our method is stable with the changes of $\lambda$ value.

\section{Conclusion}
\label{sec:conclusion}
We propose a new orthogonal perspective with respect to existing works to solve DFML problem for exploring underlying data knowledge. Our framework PURER is architecture, dataset and model-scale agnostic, thus working well in various real-world scenarios.
It contains two components: (i) ECI to perform pseudo episode training with an increasing order of difficulty during meta training; (ii) ICFIL to settle the task-distribution shift issue during meta testing. Extensive experiments show the effectiveness of PURER.

\vspace{0.15cm}

\noindent
\textbf{Acknowledgement.}\ This work is supported by the National Key R\&D Program of China (2022YFB4701400/4701402), the SZSTC Grant (JCYJ20190809172201639, WDZC20200820200655001), Shenzhen Key Laboratory (ZDSYS20210623092001004).

\clearpage
{\small
\bibliographystyle{ieee_fullname}
\bibliography{texfile/ref}
}

\clearpage

\begin{appendix}
\appendixpage
\section{Implementation Details}
\label{app:implementationdetails}
\noindent
\textbf{Implementation details for SS.}\ For SS scenario, all pre-trained models are with the same architecture. We take Conv4 
as the architecture of all pre-trained models and the meta model, the same as regular meta-learning works\cite{finn2017model,chen2020variational,liu2020adaptive,wang2019simpleshot}. Conv4 is a four-block convolution neural network, where each block consists of 32 $3 \times 3$ filters, a BatchNorm, a ReLU and an additional $2 \times 2$ max-pooling. 
The episode-batch size for each iteration is 4. We adopt the Adam optimizer to learn the dynamic dataset, base model and meta model with a learning rate of 0.25, 0.01 and 0.001, respectively. We take one-step gradient descent to perform fast adaptation for both meta training and testing. After 4000 iterations (outer loops), we add curriculum mechanism to episode inversion. At each iteration, the meta model sends a positive feedback if the sum of training accuracy over the episode batch has not increased for 6 consecutive iterations. We empirically set the scaling factor $\lambda$ as 10.  For meta testing, we calibrate the backbone for 1 iteration via Adam optimizer with a learning rate of $1e-5$. We set the temperature parameter as 0.1. Then we freeze the backbone and train a new classifier over the entire support set for 100 iterations via Adam optimizer with a learning rate of 0.01. For episode inversion, we set $\alpha_{TV}=1e-4$  and $\alpha_{l_2}=1e-5$.

\noindent
\textbf{Implementation details for SH.}\ For SH scenario, all pre-trained models are trained on the same dataset but with heterogeneous architectures. For each task, we pre-train the model with an architecture randomly selected from Conv4, ResNet-10 and ResNet-18. Compared to Conv4, ResNet-10 and ResNet-18 are larger-scale neural networks. The ResNet-10 has 4 residual stages of 1 block each which gradually decreases the spatial resolution. The ResNet-18 also has 4 residual stages but with 2 blocks per stage, which is a deeper neural network. The other configurations are the same as SS.

\noindent
\textbf{Implementation details for MH.}\ For MH scenario,  all pre-trained models are trained on multiple datasets with heterogeneous architectures. During meta training, we construct each task  from the meta training dataset randomly selected from CIFAR-FS and MiniImageNet. We pre-train the model with an architecture randomly selected from Conv4, ResNet-10 and ResNet-18. We take Conv4 as the meta model architecture. During meta testing, we report the average accuracy over 600 tasks sampled from both CIFAR-FS and MiniImageNet meta testing datasets. The other configurations are the same as SS.

\section{More Results}
\label{app:moreresults}

\noindent
\textbf{Hyperparameter Sensitivity.}\ 
We evaluate the model performance sensitivity with different values $\lambda$ in \cref{tab:sentivity}. As can be seen, the achieved performance of our method is stable with the changes of $\lambda$ value, although there are some variations among different $\lambda$ values. This advantage makes it easy to apply our method in practice.

\begin{table}[tbp]
  \centering
  \small
  \caption{Hyperparameter sensitivity on  DFML CIFAR-FS 5-way classification in SS scenario.}
    \label{tab:sentivity}
      \vspace{-0.2cm}
      \scalebox{0.99}{
  \begin{tabular}{ccc}
    \toprule

    $\boldsymbol{\lambda}$&   5-way 1-shot & 5-way 5-shot \\
    \midrule
    $\lambda$=10.0 & 38.66 $\pm$ 0.78 & 51.95 $\pm$ 0.79\\
    $\lambda$=2.0 & 37.87 $\pm$ 0.70 & 49.09 $\pm$ 0.75\\
    $\lambda$=0.5 & 38.04 $\pm$ 0.79 &48.91 $\pm$ 0.75\\
     \bottomrule
  \end{tabular}}
\end{table}

\begin{table}[]
    \centering
    \caption{Effect of the number of pre-trained models in SS scenario.}
    \begin{tabular}{ccc}
    \toprule
         \shortstack{\textbf{num}}& 5-way 1-shot & 5-way 5-shot\\
         \midrule
         2& 34.28 $\pm$ 0.75 & 45.40 $\pm$ 0.74\\
         6& 36.78 $\pm$ 0.75 & 47.71 $\pm$ 0.80 \\
         13& 38.66 $\pm$ 0.78 & 51.95 $\pm$ 0.79\\
    
    \bottomrule
    \end{tabular}
    
    \label{tab:numofpretrained}
\end{table}

\noindent
\textbf{Effect of the number of pre-trained models.} We perform the experiments about DFML with the different numbers of pre-trained models in SS scenario. We train each pre-trained model for a 5-way classification problem. \cref{tab:numofpretrained} shows the results. By increasing the number from 2 to 6, we can observe $2.5\%$ and $2.31\%$ performance gains for 1-shot and 5-shot learning, respectively. The gains increase to $4.38$ and $6.55\%$ by increasing the number from 2 to 13. The reason is that more pre-trained models can provide more underlying data knowledge of different classes. With broader data knowledge, meta-learning can acquire better generalization for target tasks with new unseen classes.

\end{appendix}

\end{document}